\documentclass[runningheads]{llncs}
\usepackage[T1]{fontenc}
\usepackage{amssymb}
\usepackage{amsmath}
\usepackage{multirow}
\usepackage{graphicx}
\usepackage{booktabs}
\usepackage{subfigure}
\newcommand{\name}{Self-Pro}
\DeclareMathOperator*{\argmin}{arg\,min}

\usepackage[misc]{ifsym}
\newcommand{\corr}{(\Letter)}
\usepackage{mwe}

\begin{document}
\tocauthor{Chenghua Gong, Xiang Li, Jianxiang Yu, Yao Cheng, Jiaqi Tan, Chengcheng Yu}
\toctitle{Self-Pro: A Self-Prompt and Tuning Framework for Graph Neural Networks}
\title{Self-Pro: A Self-Prompt and Tuning Framework for Graph Neural Networks}

\titlerunning{Self-Pro: A Self-Prompt and Tuning Framework for Graph Neural Networks}

\author{Chenghua Gong\inst{1} \and
Xiang Li \inst{1} \corr \and
Jianxiang Yu \inst{1} \and
Yao Cheng \inst{1} \and
Jiaqi Tan \inst{1} \and
Chengcheng Yu \inst{2}
}

\authorrunning{Chenghua Gong et al.}

\institute{School of Data Science and Engineering, East China Normal University, China \email{chenghuagong@stu.ecnu.edu.cn}
\email{xiangli@dase.ecnu.edu.cn}
\email{\{jianxiangyu, 52215903009, jiaqitan\}@stu.ecnu.edu.cn}
\and
School of Computer and Information Engineering, Shanghai Polytechnic University, China \email{ccyu@sspu.edu.cn}
}
\maketitle              

\begin{abstract}
    Graphs have become an important modeling tool for web applications, and Graph Neural Networks (GNNs) have achieved great success in graph representation learning. However, the performance of traditional GNNs heavily relies on a large amount of supervision.
    Recently, ``pre-train, fine-tune'' has become the paradigm to address the issues of label dependency and poor generalization.
    However, the pre-training strategies vary for graphs with homophily and heterophily, and the objectives for various downstream tasks also differ.
    This leads to a gap between pretexts and downstream tasks, resulting in ``negative transfer'' and poor performance.
    Inspired by prompt learning in Natural Language Processing (NLP), many studies turn to bridge the gap and fully leverage the pre-trained model.
    However, existing methods for graph prompting are tailored to homophily, neglecting inherent heterophily on graphs.
    Meanwhile, most of them rely on the randomly initialized prompts, which negatively impact on the stability.
    Therefore, we propose Self-Prompt, a prompting framework for graphs based on the model and data itself. 
    We first introduce asymmetric graph contrastive learning for pretext to address heterophily and align the objectives of pretext and downstream tasks.
    Then we reuse the component from pre-training phase as the self adapter and introduce self-prompts based on graph itself for task adaptation.
    Finally, we conduct extensive experiments on 11 benchmark datasets to demonstrate its superiority. We provide our codes at \url{https://github.com/gongchenghua/Self-Pro}.

\end{abstract}

\section{Introduction}
    The complex relations in the real world can be modeled using graphs. The powerful modeling capability of graphs enables the realization of web applications such as community fraud~\cite{kim2023dynamic} and automated account detection~\cite{wu2023heterophily} or drug development~\cite{chen2022drug}.
    Meanwhile, GNNs have achieved great success in representation learning of graph-structured data~\cite{kipf2016semi,velivckovic2017graph,zhu2020beyond,li2022finding}.
    However, one of the fundamental problems for traditional GNNs is that they heavily depend on task-specific labels as supervision, which are costly and difficult to obtain.
    To solve the aforementioned problems, many studies turn to ``pre-train, fine-tune'' strategies~\cite{hu2019strategies}. Through self-supervised pretexts, GNNs are pre-trained without labels and then fine-tuned with labeled data.
    However, the ``pre-train, fine-tune'' paradigm empirically faces the following challenges:

    \noindent$\bullet$ (C1) Existing pre-training methods for graph explicitly or implicitly follow the homophily assumption~\cite{liu2022revisiting}, which further restricts the generalization of GNNs on graphs with more diverse properties (e.g., heterophily). \textit{So how to find a more generalized pre-training method accommodating more diverse graphs, such as graphs with homophily and heterophily?}

    \noindent$\bullet$ (C2) The gap between objectives of the pre-training and fune-tuning phase may result in ``negative transfer''~\cite{wang2021afec} problem. For example, a typical pretext on graphs utilizes the link prediction~\cite{kipf2016variational} to map representation of adjacent nodes closer in the latent space. However, downstream task such as node classification focuses on the similarity of the intra-class and inter-class clusters. \textit{So how to reformulate the pretexts and downstream tasks into same template to avoid the negative knowledge transfer?}

    \noindent$\bullet$ (C3) During the fine-tuning phase, part or
     additional parameters of graph pre-trained models are adjusted to adapt to various downstream tasks, causing significant computational overhead. \textit{So how to fine-tune efficiently with lower cost and fully leverage the capabilities of the pre-trained model?} 

     Inspired by the success of prompt learning in NLP, ``pre-train, prompt'' has been introduced to the graph domain. The naive language prompt in NLP refers to transforming the downstream tasks into text generation template by appending prompt text since pre-trained language models have powerful text generation capabilities. For example, a sentiment analysis task for the statement ``\textit{The ECML-PKDD 2024 will accept many interesting papers!}'' can be transferred to a text generation task by appending prompt like ``\textit{The ECML-PKDD 2024 will accept many interesting papers! That is so [\_\_].}'' The language model is prone to make the prediction like ``exciting'' or ``wonderful'' rather than ``bad'' or ``sad'' in the blank. Through alignment of objectives between pretext and downstream tasks, the pretrained models can be fully leveraged and perform well even in the few-shot scenarios.
     Although some works~\cite{liu2023graphprompt,sun2022gppt,sun2023all,tan2023virtual,fang2024universal} have attempted to design the prompt framework for graphs, most of them introduce the randomly initialized parameters as virtual prompt. This greatly limits the generalization and stability of prompt learning on graphs and poses a new challenge:

     \noindent$\bullet$ (C4) Existing graph prompt methods mostly rely on the randomly initialized virtual prompt, which neglects the rich information inherently contained in the graph. \textit{So how to design prompt based on graph itself to guide the model to better leverage the pre-trained knowledge?}

     To address the aforementioned four challenges, we proposed a \underline{Self-Pro}mpt and Tuning Framework on graphs, namely, \name. Specifically, we first introduce asymmetric graph contrastive learning for pre-training since it can capture high-order information beyond local neighborhood to address heterophily. Then we reframe pretext and downstream tasks into a unified template to avoid negative knowledge transfer. Moreover, we reuse the pre-trained projector as the self-adapter for downstream, allowing for efficient tuning and fully leveraging the pre-trained knowledge. Finally, we inject self-prompt from structural and semantic perspective into downstream to provide extra guidance for task-adaptation. In summary, our main contributions can be summarized as follows:

     \noindent$\bullet$ We propose \name, a prompt and tuning framework through unified task template for GNNs. To our best knowledge, \name~is the first graph prompt framework that can accommodate various types of graphs and downstream tasks.

    \noindent$\bullet$ We propose a self-prompt strategy for \name, hinging on the self adapter, semantic and structural prompts to provide extra information and transfer the pre-trained knowledge for downstream adaptation.

    \noindent$\bullet$ We conduct extensive experiments on 11 benchmark datasets to evaluate \name. Our results show its superiority over other state-of-the-art competitors.

\section{Related Work}
    \subsection{Graph Neural Networks}
    Recently, GNNs have received significant attention for real-world applications and many advanced GNNs are proposed~\cite{kipf2016semi,velivckovic2017graph} on homophilous graphs. Most of them follow the message-passing framework, where nodes update their representations by combining and aggregating messages from their neighbors. To cope with more complex patterns, many studies have designed GNNs for heterophilous graphs~\cite{gong2024towards,li2022finding,pei2020geom,zhu2020beyond}. The core idea of GNNs for heterophily is to discriminate heterophilous neighbors and utilize multi-hop neighbors. However, these GNNs all follow the supervised learning paradigm and depend on labels heavily. To tackle this issue, many studies have shifted their focus towards unsupervised graph learning and the ``pre-train, fine-tune'' paradigm~\cite{xia2022survey}.

    \subsection{Graph Pre-training}
    Having witnessed the success of pre-training in language and vision fields, a myriad of graph pre-training methods have emerged. Graph pre-training models mine intrinsic properties on graphs through self-supervised pretexts, and then are fine-tuned for downstream tasks. 
    Graph pre-training strategies~\cite{xia2022survey} can be broadly categorized into two types: generative and contrastive methods. Generation-based methods including various Graph Auto-Encoders (GAEs)~\cite{kipf2016variational,hou2022graphmae,li2023s,li2023seegera} use the input as the supervision signal and reconstruct the feature and structure of graph. Contrast-based methods including various Graph Contrastive Learning (GCL) frameworks~\cite{velivckovic2018deep,zhu2020deep,hassani2020contrastive,yuan2023muse} construct different views for graphs through data augmentation or prior knowledge and then maximize their agreement. It is worth noting that most graph pre-training methods are exclusively applicable to graphs with homophily, only a few works~\cite{liu2023beyond,yuan2023muse,xiao2024simple} have taken into account the extensive heterophily on graphs. Meanwhile, there is an inherent gap between the objectives of pre-training and fine-tuning phase. Pretexts extract general knowledge from graphs without supervision, while downstream tasks are rooted in specific labels. This gap may have a negative impact on the knowledge transfer~\cite{wang2021afec} and hurt the performance of downstream tasks.

    \subsection{Graph Prompt Learning}
    Originated from NLP, ``pre-train, prompt'' paradigm~\cite{liu2023pre} reformulates downstream tasks into the same template of pretexts and designs prompts for downstream adaptation. Prompt learning fully unleashes the potential of pre-trained models where adjusting only a few parameters can achieve great results with limited labels. Inspired by the success of prompt learning in nature language~\cite{brown2020language,wei2021finetuned} and visual data~\cite{jia2021scaling,jia2022visual}, the graph domain has shifted the focus towards the ``pre-train, prompt'' paradigm~\cite{sun2023graph}. GPPT~\cite{sun2022gppt} and GraphPrompt~\cite{liu2023graphprompt} both align the objective of downstream tasks with link prediction. ProG~\cite{sun2023all} reformulates different-level tasks to graph level and further introduces virtual prompt graphs with meta-learning.
    GPF~\cite{fang2024universal} injects learnable prompts into the feature space for downstream task adaptation while VNT-GPPE~\cite{tan2023virtual} adjusts pre-trained graph transformers by prompting virtual nodes. Similar to graph pre-training, existing graph prompting methods only focus on homophilous graphs while overlook heterophilous graphs, which restricts their wide applications in the real word. Meanwhile, they heavily rely on the randomly initialized prompts, and the added virtual features or nodes are difficult to interpret. All of them ignore the rich information inherent in graph itself and the potential of prompt within models.

\section{Preliminaries}
    \subsubsection{Graphs.}
    Let $\mathcal{G}=(\mathcal{V}, \mathcal{E})$ be the input graph, where $\mathcal{\mathcal{V}}$ is the set of $|V|$ nodes and $\mathcal{E}$ is the set of edges. Let $\mathbf{X} \in \mathbb{R}^{|\mathcal{V}| \times D}$ be the node attribute matrix where the $i$-th row $x_i$ is the $D$-dimensional feature vector of node $v_i \in \mathcal{V}$. Graph structure can be denoted by the adjacency matrix $\mathbf{A} \in [0,1]^{|\mathcal{V}| \times |\mathcal{V}|}$  with $A_{i,j}=1$ if $e_{i,j} \in \mathcal{E}$, otherwise $A_{i,j}=0$. The neighboring set of node $v$ is denoted as $\mathcal{N}(v)$. The input graph $\mathcal{G}$ can be denoted as a tuple of matrices $G=(\mathbf{X},\mathbf{A})$.

    \subsubsection{Pre-train, Fine-tune.} The goal of graph pre-training is to learn a GNN encoder $f_{\theta}$ parameterized by $\theta$ to map each node $v$ to a low-dimensional representation $\mathbf{h}_v$ through the pretext in the pre-training phase, then $\mathbf{h}_v$ are fine-tuned according to the requirements of various downstream tasks.

    \subsubsection{Graph Contrastive Learning (GCL).} 
\begin{figure*}[htbp]
    \centering  
    \subfigure[GCL with smoothing]{
    \includegraphics[width=0.31\linewidth]{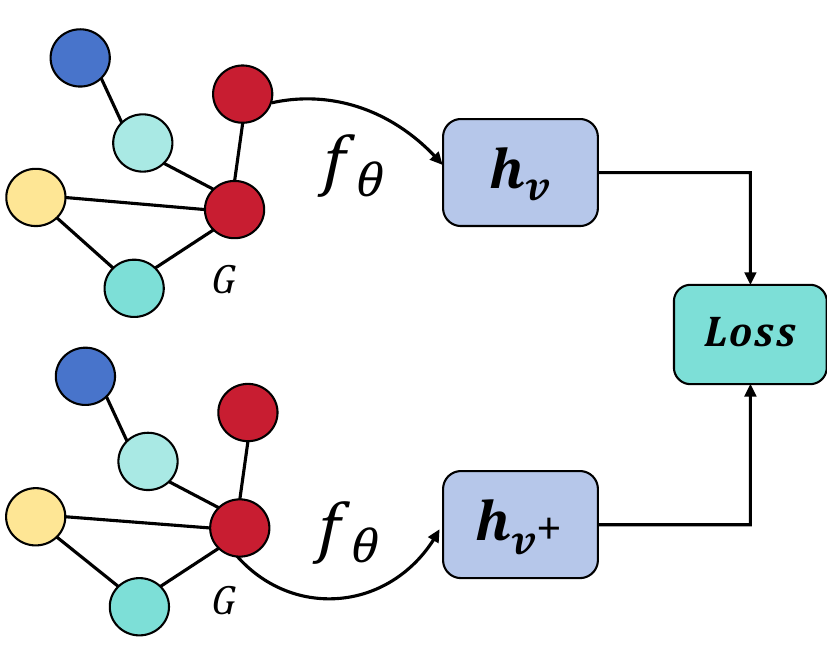}
    \label{gcl1}
    }
    \subfigure[GCL with augmentation]{
    \includegraphics[width=0.31\linewidth]{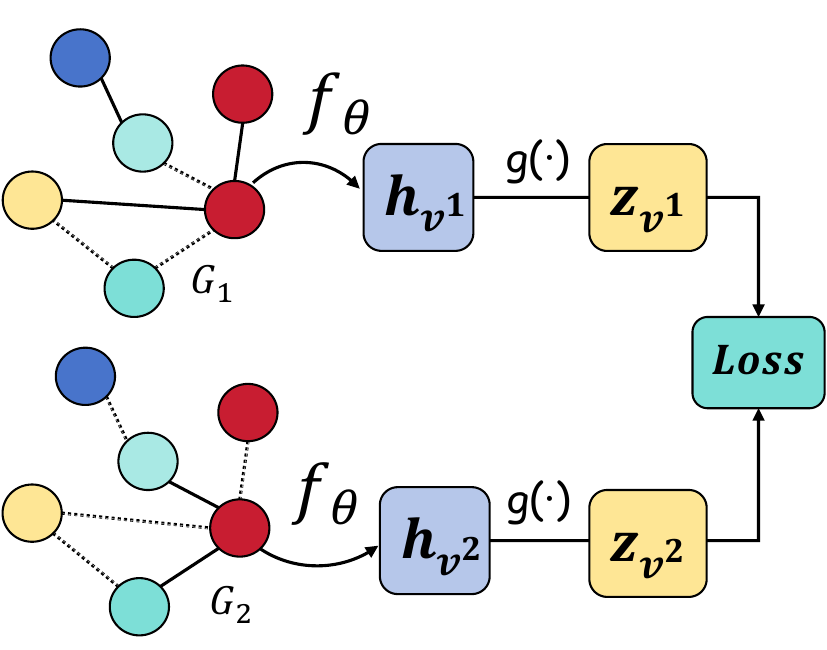}
    \label{gcl2}
    }
    \subfigure[Asymmetric GCL]{
    \includegraphics[width=0.31\linewidth]{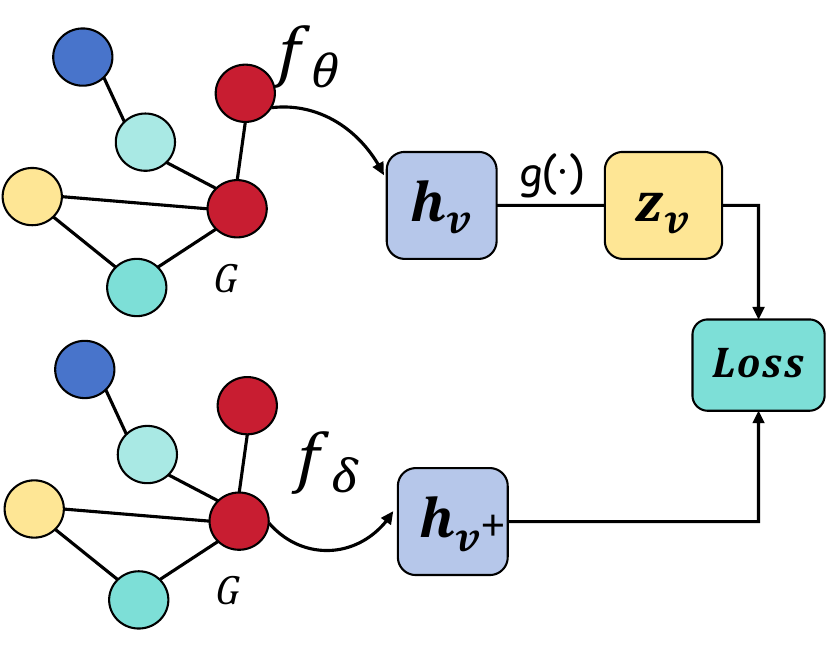}
    \label{gcl3}
    }
    \caption{Illustration of existing graph contrastive learning schemes. We primarily focus on the positive sampling process and omit the negative sampling process.}
    \end{figure*}

    GCL with smoothing~\cite{zhang2022localized} is one GCL schema which ensures that neighboring nodes the have similar representations, the objective of its pretext is:
    \begin{equation}
    \mathcal{L}=-\frac{1}{|\mathcal{V}|} \sum_{v\in\mathcal{V}} \frac{1}{|\mathcal{N}(v)|} \sum_{v^{+} \in \mathcal{N}(v)} \log 
    \frac{ \exp({\mathbf{h}_v}^\mathsf{T} \mathbf{h}_{v^{+}} / \tau)}{\exp({\mathbf{h}_v}^\mathsf{T}\mathbf{h}_{v^{+}} / \tau)+\sum\limits_{v^{-}\in\mathcal{V^{-}}}\exp({\mathbf{h}_v}^\mathsf{T}\mathbf{h}_{v^{-}} / \tau)}, 
    \end{equation}
    where $\mathbf{h}_v,\mathbf{h}_{v^{+}},\mathbf{h}_{v^{-}}$ are the representations of nodes $v,v^{+}$ and $v^{-}$ obatined by $f_{\theta}(G)$, and $\tau$ is the temperature hyper-parameter. $\mathcal{N}(v)$ is the positive sample set containing one-hop neighborhoods of node $v$ and $\mathcal{V^{-}}$ is the negative sample set randomly sampled. The illustration for this scheme of smoothing is presented in Fig.~\ref{gcl1}. 
    Another GCL schema~\cite{zhu2020deep,yuan2023muse} shown in Fig.~\ref{gcl2} learns representations by contrasting graphs under augmented views. Specifically, the representation of node $v$ in one augmented view is learned to be close to that from the other view and far away from the representations of other nodes. Given two augmented $G_1$ and $G_2$ from $G$, the objective of its pretext is:
    \begin{equation}
    \mathcal{L}=-\frac{1}{|\mathcal{V}|} \sum_{v\in\mathcal{V}} \log 
    \frac{ \exp({\mathbf{z}_{v^{1}}}^\mathsf{T} \mathbf{z}_{v^{2}} / \tau)}{\exp({\mathbf{z}_{v^{1}}}^\mathsf{T}\mathbf{z}_{v^{2}} /\tau)+\sum\limits_{v^{-}\in\mathcal{V^{-}}}\exp({\mathbf{z}_{v^{1}}}^\mathsf{T}\mathbf{z}_{v^{-}} / \tau)}. 
    \end{equation}
    Here, $\mathbf{z}_{v^{1}} = g(\mathbf{h}_{v^{1}}), \mathbf{z}_{v^{2}} = g(\mathbf{h}_{v^{2}}), \mathbf{z}_{v^{-}} = g(\mathbf{h}_{v^{-}}),$ and $g(\cdot)$ is the projector to map the representations to the latent space where the contrastive loss is applied. $\mathbf{h}_{v^{1}}$ and $\mathbf{h}_{v^{2}}$ are representations of node $v$ from two augmented views.

\section{Methods}
\begin{figure*}[htbp]
    \centering
    \includegraphics[width=0.97\linewidth]{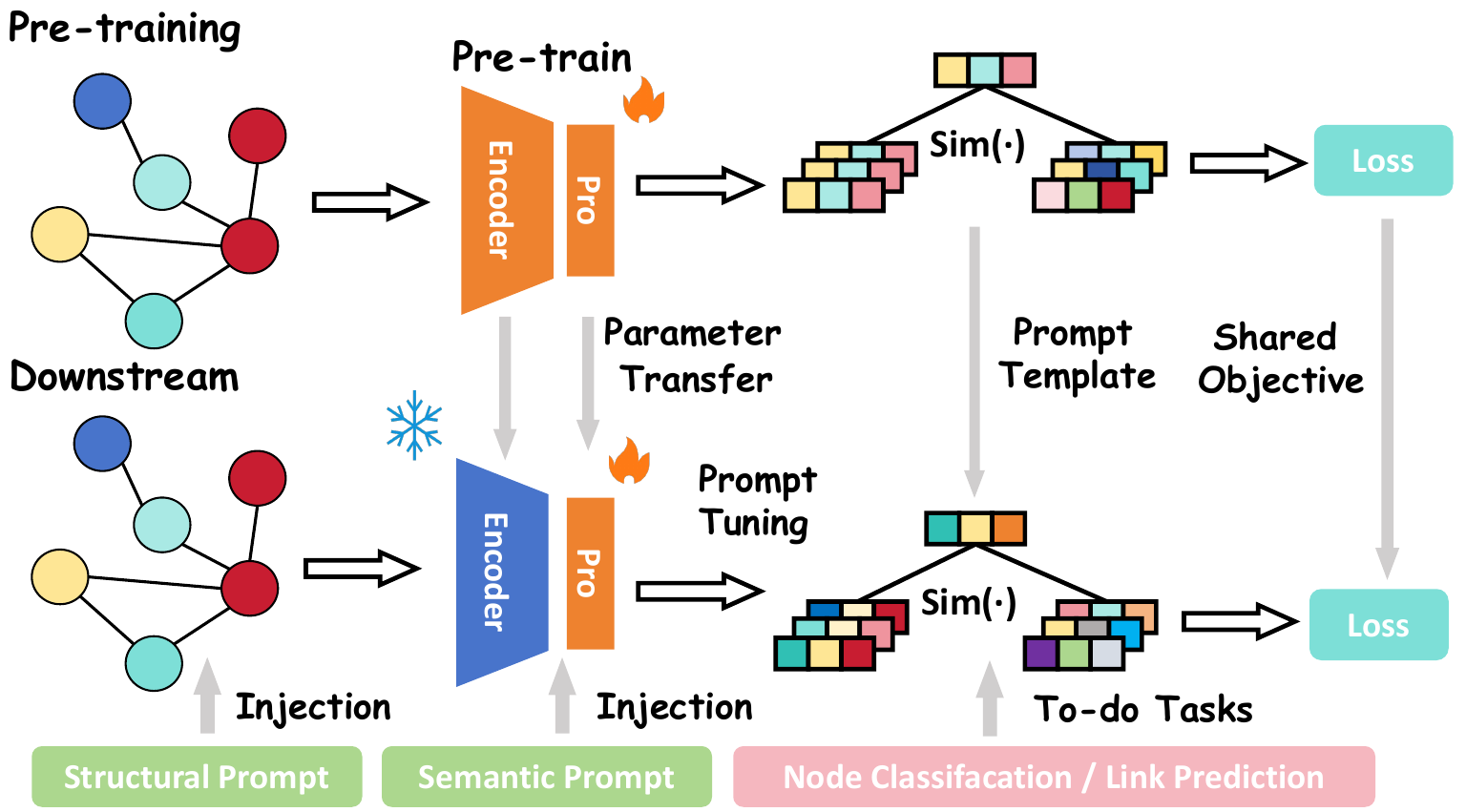}
    \caption{The overview framework of \name.}
    \label{framework}
    \end{figure*}

    In this section, we introduce the proposed framework of \name~to address the above-mentioned challenges.
    The overall framework is shown in Fig.~\ref{framework}.

    \subsection{Pre-training Method}
    For graph pre-training, contrastive methods provide broader applicability and overlapping task sub-spaces for better knowledge transfer~\cite{ma2023hetgpt}. 
    However, most GCL schemes explicitly or implicitly follow the homophily assumption~\cite{liu2022revisiting,guo2024architecture} and empirically face issues on heterophilous graphs. 
    So we choose the Graph Asymmetric Contrastive Learning (GraphACL)~\cite{xiao2024simple} as our backbone to accommodate graph with heterophily and homophily. As shown in Fig.~\ref{gcl3}, GraphACL introduces the asymmetric framework with a projector. Unlike traditional symmetric schemes, the design of projector allows neighboring nodes to have different representations, and it has been theoretically and experimentally proven in~\cite{xiao2024simple} to capture high-order neighbor similarities on graphs and effectively handle the heterophily. The pretext loss of GraphACL is:
    \begin{equation}
    \mathcal{L}=-\frac{1}{|\mathcal{V}|} \sum_{v\in\mathcal{V}} \frac{1}{|\mathcal{N}(v)|} \sum_{v^{+} \in \mathcal{N}(v)} \log 
    \frac{ \exp({\mathbf{z}_v}^\mathsf{T} \mathbf{h}_{v^+} / \tau)}{\exp({\mathbf{z}_v}^\mathsf{T} \mathbf{h}_{v^+} / \tau)+\sum\limits_{v^{-}\in\mathcal{V^{-}}}\exp({\mathbf{h}_v}^\mathsf{T}\mathbf{h}_{v^{-}} / \tau)}, 
    \label{lcon}
    \end{equation}
    where $\mathbf{z}_v = g(\mathbf{h}_v)$ and $g(\cdot)$ can be a linear transformation layer or a MLP. The positive sample set is from $\mathcal{N}(v)$ and random data augmentation are not required.
    $\mathbf{h}_{v}$ and $\mathbf{h}_{v^+}$ are obtained separately from two encoders $f_{\theta}$ and $f_{\delta}$ updated by exponential moving average to model more complex patterns and prevent $g(\cdot)$ from degenerating to the identity function~\cite{xiao2024simple}. The projector enables the model to capture higher-order context information of nodes. Similar to GCL with augmented views, GraphACL also abandons the projector and solely uses the output of encoder for downstream tasks. However, the projector also contains rich pre-trained knowledge, simply discarding it would be inappropriate. It can be combined with prompt techniques to empower the downstream tasks.

     \subsection{Prompt Template}
    Prompt learning in NLP reformulates both pretext and various downstream tasks into the unified template of masked language modeling~\cite{brown2020language,liu2023pre}. As for GCL, the core is to maximize the mutual information between positive sample pairs and minimize that between negative sample pairs. To align the pretext with downstream tasks, we focus on the most common downstream tasks on graphs: link prediction and node classification, and propose a unified prompt template, denoted as $p(\cdot)$. The template transforms various tasks into pairwise similarity calculation between tokens:
    \begin{equation}
        p(v)=sim(\mathbf{t}_v,\mathbf{t}_t),
    \end{equation}
    where $\mathbf{t}_v$ is the token of node $v$ and 
    $\mathbf{t}_t$ is the target token. Both of them can be obtained based on node representations, and $sim(\cdot)$ can be simple dot product or cosine similarity function.

    \subsubsection{Pretext.} In the pre-training phase, the targets are set as the positive and negative samples for the anchor nodes. We use the dot product in $p(\cdot)$ for simplicity and the objective of pretext is to minimize the distance between the positive sample pairs and maximize that between the negative sample pairs in the representation space as illustrated in Eq.~\ref{lcon}. 
    
    \subsubsection{Link prediction.} For link prediction, the targets are set as adjacent and non-adjacent nodes for the anchor nodes. The objective is to minimize the distance between pairs of adjacent nodes and maximize that between pairs of non-adjacent nodes in the representation space.

    \subsubsection{Node classification.} Unlike traditional node classification, we set a prototype for each class rather than predict solely by direct labels. Through class prototypes as targets, the node classification can be reformulated into pairwise similarity calculation. The objective is to minimize the distance between nodes and the intra-class prototype, while maximize that between nodes and the inter-class prototype. Formally, we denote the set of class prototype tokens as $\mathcal{C}= \{ \mathbf{t}_1,\mathbf{t}_2,\dots, \mathbf{t}_C \}$, where $C$ is the number of classes. Each prototype token shares the same dimension with the node token and preserves class-specific semantics. Each prototype is initialized by the mean token of labeled nodes for each class~\cite{sun2022gppt,liu2023graphprompt,ma2023hetgpt}:
    \begin{equation}
        \mathbf{t}_c = \frac{1}{N_c} \sum_{v \in \mathcal{V}_L,y_v=c} \mathbf{t}_v, \forall c \in {1,2,\dots C},
    \end{equation}
    where $N_c$ is the number of nodes with class $c$ in the labeled set $\mathcal{V}_L$ and $\mathbf{t}_v$ denotes the token of node $v$. Through this initialization, node classification can be aligned with common patterns for afterward prompt tuning. 
    
    \subsection{Self-Prompt Architecture}
    Most ``pre-train,fine-tune'' approaches for graphs introduce an additional layer for the downstream fine-tuning. Meanwhile, the projector in GCL plays a crucial role~\cite{guo2024architecture} in the pretext and contains partial knowledge obtained during pre-training. For example, the projector in GraphACL~\cite{xiao2024simple} is to preserve the context information of nodes and model local neighborhood distribution beyond homophily assumption. However, the projector is always discarded after pre-training. This motivates us to further reuse the projector as the self adapter and construct a naturally parameterized architecture for tuning. Given the input graph $\mathcal{G}$, an encoder $f_{\theta}$ parameterized by $\theta$ and the projector $g_{\phi}$ parameterized by $\phi$, the pre-training objective can be formally formulated as:
    \begin{equation}
        \theta^{*}, \phi^{*} = \argmin_{\theta,\phi} \mathcal{L}_{pre}(f_{\theta}, g_{\phi}, \mathcal{G}),
    \end{equation}
    where $\mathcal{L}_{pre}$ is the loss of pretext illustrated in Eq.~\ref{lcon}. For downstream tasks, we freeze the pre-trained GNN encoder $f_{\theta}$ and only optimize the parameters of projector $g_{\phi}$. Given a set of labeled nodes $\mathcal{V}_L$ and corresponding label set $\mathcal{Y}$, the objective for the downstream tasks can be denoted as:
    \begin{equation}
        \phi^{**} = \argmin_{\phi^{*}} \mathcal{L}_{dow}(g_{\phi^{*}}, \mathcal{V}_L, \mathcal{Y}),
    \end{equation}
    where $\mathcal{L}_{dow}$ is optimized by the supervision and will be introduced in Section~\ref{tuning}. Note the parameters of projector $g_{\phi}$ are initialized from pre-training stage, denoted as $g_{\phi}^{*}$. By reusing the projector as a self adapter for downstream tasks, the pre-trained knowledge can be retained and smoothly transferred to the downstream, greatly improving the efficiency of tuning for specific tasks without additional parameters. 

    \subsection{Self-Prompt Injection and Tuning}
    \label{tuning}
    Although the objectives of pretext and downstream tasks are aligned, task-specific prompt is still required for downstream adaptation. Currently, most works introduce trainable vectors with random initialization for prompting. The specific forms include injecting prompts in the feature space~\cite{fang2024universal} and representation space~\cite{liu2023graphprompt}, or using virtual nodes~\cite{tan2023virtual} or graphs~\cite{sun2023all} as prompts. This strategy is not stable and relies on the initialization, motivating us to design more stable prompts. Considering two key elements of the graph, namely attributes and structure, we approach the problem from these two perspectives and design prompts based on the rich information inherent in the graph itself.
    
    \subsubsection{Self structural prompt.} Instead of introducing additional parameters, the original input can be modified to provide self prompts. From structural perspective, we can add or remove edges on the adjacency matrix based on prior knowledge for prompt. There have been studies~\cite{zhu2020beyond,li2022finding} emphasizing the importance of multi-hop neighbors on graphs. Especially, most nodes share the same label with their two-hop neighbors on most homophilous graphs~\cite{xiao2024simple}. Therefore, we can simply add two-hop neighbors into the adjacency matrix as self structural prompt for node classification. Formally, we define the two-hop graph $\mathcal{G}_{2}$ from $\mathcal{G}$ and its adjacency matrix is $\mathbf{A}_2$, where $(\mathbf{A}_2)_{ik}=1$ if there exists $j$ such that $e_{ij} \in \mathcal{E}$ and $e_{jk} \in \mathcal{E}$. Then we feed the $\mathcal{G}_{2}$ into our frozen pre-trained encoder $f_{\theta}$ to obtain token $\mathbf{t}_v$ of node $v$ for tuning:  
    \begin{equation}
        \mathbf{t}_v = f_{\theta}(\mathcal{G}_{2})[v] = f_{\theta}(\mathbf{A}_2, \mathbf{X})[v].
    \end{equation}
    It is worth noting that there are many available methods for self structural prompts, such as assigning edge weights or using other structural learning techniques~\cite{zhao2023graphglow,gong2024towards} to obtain new structural prompt. However, we inject two-hop neighbors as prompts in our implementation for simplicity.

    \subsubsection{Self semantic prompt.} Compared with graph structure, node attributes can be even more important in certain situations, such as on heterophilous graph. But how to transform the semantic of attributes into the representation space is a challenge. Inspired by~\cite{yuan2023muse}, we construct a unit matrix $\mathbf{I} \in \mathbb{R}^{|\mathcal{V}| \times |\mathcal{V}|}$ to replace the adjacency matrix $\mathbf{A}$ of the original graph $\mathcal{G}$. Then we feed the prompt graph $\mathcal{G}_I$ into the pre-trained encoder $f_{\theta}$ to obtain semantic token $\mathbf{s}_v$:
    \begin{equation}
        \mathbf{s}_v = f_{\theta}(\mathcal{G}_{I})[v] = f_{\theta}(\mathbf{I}, \mathbf{X})[v].
    \end{equation}
    Replacing the adjacency matrix with an identity matrix can preserve the semantic information of each node as much as possible. Then we inject the self semantic prompt into the contextual information $\mathbf{h}_v$ of the node $v$, where
    \begin{equation}
        \mathbf{h}_v = f_{\theta}(\mathcal{G})[v] = f_{\theta}(\mathbf{A}, \mathbf{X})[v].
    \end{equation}
    To obtain the node token $\mathbf{t}_v$ for tuning, we can adopt many injection patterns for semantic prompts $\mathbf{s}_v$ and contextual information $\mathbf{h}_v$. One way is to set a hyper-parameter $\mu$ to adjust the weights of the two components:
    \begin{equation}
        \mathbf{t}_v = \mu \mathbf{s}_v + (1-\mu) \mathbf{h}_v.
    \end{equation}
    Moreover, we can adopt a more fine-grained strategy and set a self weight $w_v$ for node $v$ to model the injection proportion:
    \begin{equation}
        \mathbf{t}_v = w_v  \mathbf{s}_v+ (1 - w_v) \mathbf{h}_v, w_v = sim(h_v,s_v),
    \end{equation}
    where we can reuse the similarity function $sim(\cdot)$ in prompt template $p(\cdot)$.
    
    \subsubsection{Prompt tuning.} After prompt injection, we feed $\mathbf{t}_v$ into the adapter $g_{\phi}$ for integration and obtain prompted token $\mathbf{t'}_v = g_{\phi}(\mathbf{t}_v)$ for tuning. For node classification, we follow the prompt template $p(\cdot)$ and same objective of the pretext:
    \begin{equation}
        \mathcal{L}_{dow}=-\sum_{v\in \mathcal{V}_L} \log \frac{\exp(\mathbf{t'}_v^{\mathsf{T}} \mathbf{t'}_{y_v} / \tau) }{\exp(\mathbf{t'}_v^{\mathsf{T}} \mathbf{t'}_{y_v} / \tau) + \sum_{c=1, c \neq y_v}^{C}\exp(\mathbf{t'}_v^{\mathsf{T}} \mathbf{t'}_{c} / \tau)},
    \end{equation}
    where $\mathbf{t'}_{y_v}$ is the token of intra-class prototype for $v$ and the label of node to predict is consistent with the nearest class prototype.
    For link prediction, we randomly sample one positive node $a$ from adjacent neighbors of $v$, and one negative node $b$ that does not directly link to $v$, forming a triplet $(v,a,b)\in \mathcal{T}$. Then the downstream objective is given as:
    \begin{equation}
        \mathcal{L}_{dow}=-\sum_{(v,a,b) \in \mathcal{T}} \log \frac{\exp(\mathbf{t'}_v^{\mathsf{T}} \mathbf{t'}_a / \tau) }{\exp(\mathbf{t'}_v^{\mathsf{T}} \mathbf{t'}_a / \tau) + \exp(\mathbf{t'}_v^{\mathsf{T}} \mathbf{t'}_b / \tau)}. 
    \end{equation}
    
\section{Experiments}
    In this section, we perform experiments on benchmark datasets to evaluate \name~and attempt to answer the following questions:

    \noindent$\bullet$(RQ1) How effective and efficient is \name~for node classification on homophilous and heterophilous graphs under the few-shot setting?
    
    \noindent$\bullet$(RQ2) How adaptable is \name~when transferred to different shots and tasks?
    
    \noindent$\bullet$(RQ3) How does \name~perform under different prompts and without prompt?
    
    \noindent$\bullet$(RQ4) How does hyper-parameters influence the performance of \name?
    
    \subsection{Experimental Settings}
    \subsubsection{Datasets.} We conduct experiments on homophilous and heterophilous graphs. For graphs with homophily, we adopt citation networks: Cora, Citeseer and Pubmed~\cite{kipf2016semi}, and co-purchase graphs: Computer and Photo~\cite{kipf2016semi}. For graphs with heterophily, we adopt webpage networks: Texas, Wisconsin, Cornell~\cite{pei2020geom}, and Wikipedia page–page networks: Chameleon, Squirrel and Crocodile~\cite{rozemberczki2021multi}. Further details of the datasets are summarized in Table~\ref{datasets}.
    \begin{table}[!htbp]
    \centering
    \caption{Statistics of datasets used in experiments.}
    \label{datasets}
    \begin{tabular}{c c c c c c c}
    \toprule
    \textbf{Dataset} & \textbf{Classes} & \textbf{Nodes} & \textbf{Edges} & \textbf{Features} \\
    \midrule
    Cora & 7 & 2,708 & 5,429 & 1,433 \\
    Citeseer & 6 & 3,327 & 4,732 & 3,703 \\
    Pubmed & 3 & 19,717 & 44,338 & 500 \\
    Photo & 8 & 7,650 & 119,081 & 745 \\
    Computer & 10 & 13,752 & 574,418 & 767 \\
    \midrule
    Texas & 5 & 183 & 309 & 1,703 \\
    Cornell & 5 & 183 & 295 & 1,703 \\
    Wisconsin & 5 & 251 & 499 & 1,703 \\
    Chameleon & 5 & 2,277 & 36,101 & 2,325 \\
    Squirrel & 5 & 5,201 & 216,933 & 2,089 \\
    Crocodile & 5 & 11,631 & 360,040 & 2,089 \\
    \bottomrule
    \end{tabular}
    \end{table}
    \begin{table}[h]
    \centering
    \caption{Node classification accuracy ($\%$) on homophilous graphs. The best and second best performance for each dataset are marked with bold and underline, respectively.}
    \label{homophily node}
    \begin{tabular}{ c | c | c | c | c | c }
    \toprule
    \textbf{Method} & \textbf{Cora} & \textbf{Citeseer} & \textbf{Pubmed} & \textbf{Computer} & \textbf{Photo}\\ 
    \midrule
    GCN & 41.04\tiny{±6.06} & 33.83\tiny{±6.53} & 55.83\tiny{±7.59} & 46.93\tiny{±8.85} & 58.62\tiny{±7.85}\\
    GAT & 44.39\tiny{±7.31} & 36.72\tiny{±5.82} & 56.74\tiny{±9.12} & 48.63\tiny{±7.93} & 57.15\tiny{±5.42}\\
    \midrule
    GAE & 47.51\tiny{±8.53} & 40.68\tiny{±7.23} & 53.72\tiny{±6.92} & 50.53\tiny{±11.05} & 62.73\tiny{±7.29}\\
    DGI & 54.11\tiny{±9.72} & 45.19\tiny{±9.14} & 54.36\tiny{±10.03} & 52.38\tiny{±9.57} & 64.38\tiny{±8.13}\\
    MVGRL & 56.02\tiny{±10.86} & \underline{46.25\tiny{±9.98}} & 56.29\tiny{±8.53} & 53.14\tiny{±10.71} & 64.52\tiny{±6.67}\\
    GRACE & 55.56\tiny{±9.76} & 45.34\tiny{±11.52} & 54.16\tiny{±9.05} & 52.72\tiny{±9.98} & 62.31\tiny{±7.94}\\
    MUSE & 54.13\tiny{±10.21} & 43.81\tiny{±7.71} & 53.85\tiny{±8.69} & 50.61\tiny{±11.52} & 62.78\tiny{±9.58}\\
    GraphACL & 56.56\tiny{±11.34} & 45.64\tiny{±9.71} & \underline{56.73\tiny{±9.73}} & 53.25\tiny{±10.17} & \underline{69.66\tiny{±7.23}}\\
    \midrule
    GPPT & 41.63\tiny{±8.76} & 42.89\tiny{±7.93} & 50.98\tiny{±9.12} & 39.48\tiny{±12.08} & 53.73\tiny{±8.12}\\
    GraphPrompt & 55.32\tiny{±10.56} & 45.84\tiny{±11.61} & 53.49\tiny{±11.18} & 45.32\tiny{±8.87} & 61.32\tiny{±6.85}\\
    ProG & \underline{56.82\tiny{±10.53}} & 45.72\tiny{±10.25} & 56.23\tiny{±9.81} & \underline{53.32\tiny{±9.31}} & 68.09\tiny{±7.01}\\
    \name & \textbf{60.63\tiny{±11.03}} & \textbf{48.48\tiny{±9.85}} & \textbf{60.10\tiny{±9.16}} & \textbf{57.12\tiny{±9.06}} & \textbf{75.90\tiny{±6.74}} \\
    \bottomrule
    \end{tabular}
    \end{table}
    \subsubsection{Setup and evaluation.} 
    For node classification, we follow the $k$-shot setup in~\cite{liu2023graphprompt}. Specifically, we randomly sample 1 node per class for training, 5 nodes per class for validation and the remaining nodes for testing. We run the experiments 10 times, and report mean accuracy with standard deviation.
    For link prediction, we follow the setup in~\cite{li2023seegera} and construct the validation/test set by randomly selecting 20\%/10\% edges from the original graph. After removing the selected edges, we train the model with the remaining 70\% edges and use the dot-product decoder. We run the experiments 10 times and  use two commonly used metrics, the area under the ROC curve (AUC) and the average precision (AP), to report the average results with standard deviation.  
    
    \subsubsection{Baselines.} To evaluate \name, we compare it with the state-of-the-art methods from three categories:
    
    \noindent$\bullet$ Supervised methods: GCN~\cite{kipf2016semi}, GAT~\cite{velivckovic2017graph}, H$_2$GCN~\cite{zhu2020beyond}, GloGNN~\cite{li2022finding}, MLP.

    \noindent$\bullet$ Graph pre-training methods: GAE~\cite{kipf2016variational}, DGI~\cite{velivckovic2018deep}, GRACE~\cite{zhu2020deep}, MVGRL~\cite{hassani2020contrastive}, MaskGAE~\cite{li2023s}, 
    SeeGera~\cite{li2023seegera},
    MUSE~\cite{yuan2023muse}, GraphACL~\cite{xiao2024simple}.
    
    \noindent$\bullet$ Graph prompt methods: GPPT~\cite{sun2022gppt}, GraphPrompt~\cite{liu2023graphprompt} and ProG~\cite{sun2023all}.
    
    \begin{table}[h]
    \caption{Node classification accuracy ($\%$) on heterophilous graphs. The best and second best performance for each dataset are marked with bold and underline.}
    \label{heterophily node}
    \begin{tabular}{c | c | c | c | c | c | c }
    \toprule
    \textbf{Method} & \textbf{Texas} & \textbf{Wisconsin} & \textbf{Cornell} & \textbf{Chameleon} & \textbf{Squirrel} & \textbf{Crocodile} \\
    \midrule
    MLP & \underline{38.74\tiny{±10.72}} & 36.51\tiny{±11.45} & 32.61\tiny{±11.34} & 22.67\tiny{±2.88} & 16.28\tiny{±2.52} & 20.72\tiny{±4.98} \\
    GCN & 19.48\tiny{±11.17} & 19.84\tiny{±13.67} & 23.93\tiny{±9.33} & 26.84\tiny{±7.52} & 17.34\tiny{±2.46} & 20.08\tiny{±5.45}\\
    GAT & 20.40\tiny{±12.13} & 18.27\tiny{±10.19} & 23.77\tiny{±10.25} & 25.75\tiny{±5.46} & 16.63\tiny{±2.07} & 19.96\tiny{±4.58}\\
    H$_2$GCN & 34.37\tiny{±15.89} & \underline{37.03\tiny{±12.53}} & 28.38\tiny{±12.52} & 26.95\tiny{±3.73} & 19.21\tiny{±3.19} & 24.37\tiny{±5.79}\\
    GloGNN & 35.21\tiny{±16.47} & 35.12\tiny{±13.15} & 30.72\tiny{±14.68} & 26.82\tiny{±3.63} & 20.84\tiny{±3.45} & 24.02\tiny{±6.53} \\
    
    \midrule
    GAE & 21.89\tiny{±16.93} & 17.71\tiny{±8.79} & 21.44\tiny{±8.97} & 22.80\tiny{±3.63} & 18.21\tiny{±3.17} & 22.83\tiny{±6.23}\\
    DGI & 19.21\tiny{±12.71} & 20.64\tiny{±11.89} & 21.35\tiny{±9.74} & 23.42\tiny{±4.82} & 17.83\tiny{±2.34} & 24.90\tiny{±7.04}\\
    GRACE & 18.58\tiny{±9.38} & 19.95\tiny{±9.94} & 19.51\tiny{±8.07} & 24.68\tiny{±3.79} & 15.79\tiny{±1.86} & 25.02\tiny{±6.48} \\
    MUSE & 22.47\tiny{±10.74} & 24.91\tiny{±6.17} & 32.16\tiny{±14.79} & 27.72\tiny{±3.75} & \underline{21.53\tiny{±2.83}} & 28.73\tiny{±5.95}\\
    GraphACL & 27.56\tiny{±15.21} & 31.84\tiny{±16.18} & \underline{37.15\tiny{±12.92}} & \underline{27.94\tiny{±4.89}} & 20.02\tiny{±2.45} & \underline{32.88\tiny{±5.54}} \\
    
    \midrule
    GPPT & 16.96\tiny{±12.69} & 14.62\tiny{±10.77} & 17.57\tiny{±9.13} & 23.05\tiny{±3.68} & 17.84\tiny{±2.45} & 20.83\tiny{±4.73}\\
    GraphPrompt & 23.13\tiny{±11.84} & 28.54\tiny{±7.64} & 28.67\tiny{±16.24} & 25.62\tiny{±4.66} & 18.02\tiny{±2.78} & 21.58\tiny{±7.12}\\
    ProG & 26.81\tiny{±14.27} & 23.09\tiny{±9.49} & 26.82\tiny{±13.68} & 22.73\tiny{±3.97} & 19.17\tiny{±2.45} & 23.95\tiny{±5.86}\\
    \name & \textbf{47.48\tiny{±15.87}} & \textbf{43.54\tiny{±12.22}} & \textbf{47.81\tiny{±14.38}} & \textbf{29.02\tiny{±4.24}} & \textbf{24.36\tiny{±3.14}} & \textbf{35.81\tiny{±5.21}}\\
    \bottomrule
    \end{tabular}
    \end{table}
    
    \subsection{Few-shot Node Classification (RQ1)}
    For node classification, we inject either structural or semantic prompt separately and report the best results. More experiments and analysis about different prompts will be elaborated on in Section~\ref{ablation}. 
    
    \subsubsection{Few-shot on homophilous graphs.} Table~\ref{homophily node} illustrates the results of few-shot node classification on homophilous graphs. These observations are as following:\\
    (1) Supervised GNN models achieve poor performance in most cases, demonstrating heavy dependency on labeled data. However, some pre-trained models achieve even worse performance than supervised models on Pubmed, indicating the negative transfer of pre-trained knowledge.\\
    (2) \name~performs the best across all the datasets and other prompt methods such as ProG also achieve good results in few-shot setting, demonstrating that a unified task template can fully leverage the capabilities of pre-trained models.
    \subsubsection{Few-shot on heterophilous graphs.}We also present the results of few-shot node classification on heterophilous graphs in Table~\ref{heterophily node}. These observations are as following:\\
    (3) Supervised methods such as H$_2$GCN, GloGNN and self-supervised methods such as MUSE and GraphACL designed for heterophily achieve good results due to the ability to capture higher-order similarities beyond local neighborhood. However, GAE, GraphPrompt, DGI and GRACE achieve poor performance on heterophilous datasets since they follow the homophily assumption.\\
    (4) MLP achieves good results on Cornell, Texas and Wisconsin, indicating that the semantic of node attributes plays a vital role on heterophilous graph.\\
    (5) \name~achieves the best performance across all datasets and significantly outperforms GraphACL, demonstrating the superiority of the unified task template and self prompt tuning.\\
    \begin{table}[h]
    \centering
    \caption{Study of parameter efficiency on node classification.}
    \label{parameter efficiency}
    \begin{tabular}{c c c c c c}
    \toprule
    \textbf{Method} & \textbf{Cora} & \textbf{Citeseer} & \textbf{Pubmed} & \textbf{Computer} & \textbf{Photo}\\
    \midrule
    MVGRL & 3,591 & 3,078 & 1,539 & 4,104 & 5,130\\
    GRACE & 1,799 & 1,542 & 1,028 & 2,056 & 2,570\\
    GraphPrompt & 256 & 256 & 256 & 256 & 256\\
    \name & 0 & 0 & 0 & 0 & 0\\
    \bottomrule
    \end{tabular}
    \end{table}

    \subsubsection{Parameter efficiency.} 
    We also compare the number of parameters to be introduced in \name~for node classification task with a few representative models in Table~\ref{parameter efficiency}. GRACE and MVGRL employ a linear classifier for downstream tuning, while GraphPrompt introduces a prompt vector which shares the same dimension with the node representation for prompt tuning.
    For proposed \name, it outperforms the baselines GRACE, MVGRL and GraphPrompt without additional parameters as we have seen earlier, demonstrating its superiority.
    \begin{figure}[htbp]
    \centering
    \subfigure[Pubmed]{
    \includegraphics[width=0.38\linewidth]{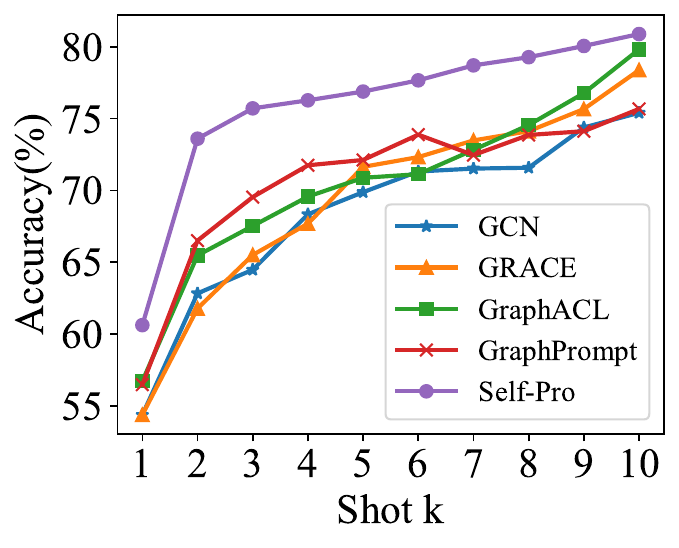}
    }
    \subfigure[Chameleon]{
    \includegraphics[width=0.38\linewidth]{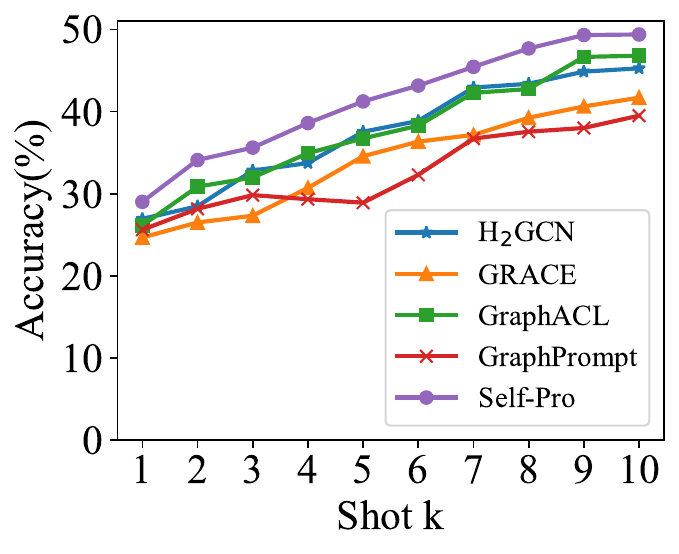}
    }
    \caption{Impact of different shots on few-shot classification in \name.}
    \label{shotnumber}
    \end{figure}
    \begin{table}
    \centering
    \caption{Node classification accuracy ($\%$) under semi-supervised setting. The best and second best performance for each dataset are marked with bold and underline.}
    \label{semisupervised}
    \begin{tabular}{ c | c | c | c | c | c }
    \toprule
    \textbf{Method} & \textbf{Cora} & \textbf{Citeseer} & \textbf{Pubmed} & \textbf{Texas} & \textbf{Cornell}\\ 
    \midrule
    MLP & 56.11\tiny{±0.34} & 56.91\tiny{±0.46} &  71.35\tiny{±0.05} & 81.62\tiny{±3.51} & 81.08\tiny{±0.98}\\
    GCN & 81.58\tiny{±0.12} & 70.30\tiny{±0.21} & 78.97\tiny{±0.35} & 60.00\tiny{±2.85} & 57.03\tiny{±2.30}\\
    GAT & 83.03\tiny{±0.24} & 72.37\tiny{±0.42} & 79.03\tiny{±0.32} & 61.62\tiny{±3.87} & 78.63\tiny{±5.92}\\
    DGI & 82.37\tiny{±0.62} & 71.94\tiny{±0.78} & 79.43\tiny{±0.84} & 60.59\tiny{±2.56} & 63.35\tiny{±1.61}\\
    GRACE & 83.01\tiny{±0.53} & 72.41\tiny{±0.38} &  81.15\tiny{±0.34} & 58.57\tiny{±2.68} & 55.86\tiny{±1.95} \\
    MVGRL & 83.03\tiny{±0.37} & 72.84\tiny{±0.64} &  79.95\tiny{±0.38} & 63.83\tiny{±5.16} & 64.08\tiny{±5.73} \\
    MUSE & 82.24\tiny{±0.47} & 71.14\tiny{±0.82} & 81.90\tiny{±0.59} & \underline{83.73\tiny{±1.39}} & \underline{82.09\tiny{±2.58}}\\
    GraphACL & \underline{84.20\tiny{±0.31}} & \textbf{73.58\tiny{±0.41}} & \textbf{82.20\tiny{±0.73}} & 71.08\tiny{±0.57} & 59.95\tiny{±2.23}\\
    \name & \textbf{84.58\tiny{±0.18}} & \underline{73.15\tiny{±0.68}} & \underline{82.11\tiny{±0.59}} & \textbf{85.14\tiny{±0.63}} & \textbf{86.29\tiny{±0.87}} \\
    \bottomrule
    \end{tabular}
    \end{table}
\subsection{Performance under different shots and tasks (RQ2)}
    \subsubsection{Performance under different shots.} We study the impact of shot number on the Pubmed and Chameleon in Fig.~\ref{shotnumber}. We vary the number of shots between 1 and 10, and compare with competitive baselines. Note that 10 shots per class might be sufficient for semi-supervised node classification. We can see that \name~consistently outperforms the baselines on Pubmed and Chameleon. 
    Further, we investigate the performance of \name~under the semi-supervised setting. For homophilous graphs, we adopt the public splits with 20 nodes per class for training, 500 nodes for validation and 1,000 nodes for testing~\cite{kipf2016semi,yuan2023muse}. For heterophilous graphs, we adopt the commonly used training/validation/test split ratio of 48/32/20 as previous works~\cite{pei2020geom,yuan2023muse}. The results are shown in Table~\ref{semisupervised} and we can find that:\\
    (1) When labels are sufficient, GraphACL and MUSE demonstrate impressive performance on both homophilous and heterophilous graphs. This can be attributed to their frameworks that adequately consider the heterophily on graphs.\\
    (2) \name~still outperforms traditional supervised models and self-supervised GNN models across most datesets, indicating that our framework can effectively transfer the pre-trained knowledge to downstream tasks once again.\\
    (3) Most baselines perform even worse than simple MLP on heterophilous graphs, indicating the importance of feature semantic. This also confirms that our model can handle the heterophily on graphs through semantic prompt injection.
    \begin{table}[h]
    \centering
    \caption{Link prediction results on homophilous graphs. The best and second best performance for each dataset are marked with bold and underline.}
    \label{link prediction}
    \begin{tabular}{ c c c c c c c}
    \toprule
    \multicolumn{1}{l}{\textbf{Metric}} & \textbf{Method} & \textbf{Cora} & \textbf{Citeseer} & \textbf{Pubmed} & \textbf{Computer} & \textbf{Photo}\\
    \toprule
    \multirow{9}{*}{AUC}       
    & GAE &94.65\tiny{±0.21}  &95.02\tiny{±0.51} &95.87\tiny{±0.84} &92.57\tiny{±0.43} &95.51\tiny{±0.82}\\
    & MaskGAE &95.45\tiny{±0.26} &\underline{97.12\tiny{±0.27}} &\underline{96.71\tiny{±0.78}} &78.53\tiny{±3.07} &85.32\tiny{±0.54}\\
    & SeeGera &\underline{95.46\tiny{±0.72}} & 94.63\tiny{±0.75} &95.39\tiny{±1.79} &96.51\tiny{±0.47} &\underline{98.34\tiny{±0.52}}\\
    & MVGRL &93.19\tiny{±0.62} &89.15\tiny{±5.13} &95.02\tiny{±0.82} &91.79\tiny{±0.82} &70.81\tiny{±3.61}\\
    & GRACE &87.29\tiny{±0.36} &87.64\tiny{±0.96} &94.39\tiny{±0.85} &84.34\tiny{±0.67} &82.78\tiny{±0.72}\\
    & GraphACL &94.88\tiny{±0.71} &95.07\tiny{±0.82} &96.12\tiny{±0.97} &89.42\tiny{±2.01} &93.19\tiny{±0.94}\\
    & GraphPrompt &90.40\tiny{±0.47} &93.83\tiny{±0.45} &91.63\tiny{±0.78} &88.57\tiny{±0.43} &85.51\tiny{±0.82}\\
    & ProG &93.79\tiny{±1.29} &93.98\tiny{±0.89} &95.92\tiny{±0.36} &91.34\tiny{±0.82} &88.97\tiny{±0.37}\\
    & Self-Pro  &\textbf{95.48\tiny{±0.15}} & \textbf{97.22\tiny{±0.30}} &\textbf{96.95\tiny{±0.65}} &\textbf{97.82\tiny{±0.32}} &\textbf{98.68\tiny{±0.48}}\\
    \midrule  
    \multirow{9}{*}{AP}       
    & GAE &94.38\tiny{±0.18} &95.65\tiny{±0.43} &95.09\tiny{±0.28} &90.07\tiny{±0.28} & 94.85\tiny{±0.45}\\
    & MaskGAE &94.67\tiny{±0.23} &96.83\tiny{±0.42} &\underline{96.89\tiny{±0.20}} &76.51\tiny{±3.92} &79.11\tiny{±0.47}\\
    & SeeGera &\textbf{95.90\tiny{±0.63}} &\textbf{97.28\tiny{±0.89}} &95.64\tiny{±1.05} &\underline{96.21\tiny{±0.36}} &\underline{98.12\tiny{±0.56}}\\
    & MVGRL &92.81\tiny{±0.87} &89.37\tiny{±4.26} &95.37\tiny{±0.21} &91.74\tiny{±0.40} &65.79\tiny{±0.46}\\
    & GRACE &85.73\tiny{±0.43} &84.95\tiny{±0.40} &93.83\tiny{±1.03} &85.71\tiny{±0.83} &81.18\tiny{±0.37}\\
    & GraphACL &94.77\tiny{±0.56} &93.87\tiny{±0.71} &96.35\tiny{±0.74} &89.28\tiny{±1.78} &92.65\tiny{±0.69}\\
    & GraphPrompt &90.67\tiny{±0.39} &93.21\tiny{±0.48} &91.70\tiny{±0.31} &89.07\tiny{±0.28} &84.85\tiny{±0.62}\\
    & ProG &93.55\tiny{±1.41} &96.01\tiny{±0.79} &95.56\tiny{±0.29} &92.04\tiny{±0.50} &89.57\tiny{±0.75}\\
    & Self-Pro  &\underline{95.74\tiny{±0.40}} &\underline{96.94\tiny{±0.21}} &\textbf{97.11\tiny{±0.41}} &\textbf{97.91\tiny{±0.74}} &\textbf{98.71\tiny{±0.31}}\\
    \bottomrule 
    \end{tabular}
    \end{table}
    \subsubsection{Performance on link prediction.}
    Table~\ref{link prediction} list the results of link prediction and show the performance of Self-Pro when transferred to different downstream tasks. We have the following findings:\\
    (1) Generative methods such as SeeGera and MaskGAE perform better than contrastive methods since their objective is to reconstruct the graph structure.\\
    (2) Although GraphPrompt uses link prediction as pretext, the poor performance could be attributed to the instability caused by the randomly initialized prompts.\\
    (3) Self-Pro even outperforms state-of-the-art generative models such as SeeGera and MaskGAE on most datasets due to better adaptability to downstream tasks through prompt tuning.
    
\subsection{Ablation Study (RQ3)}
    \label{ablation}
\begin{figure}[htbp]
    \centering
    \subfigure[Cora]{
    \includegraphics[width=0.31\linewidth]{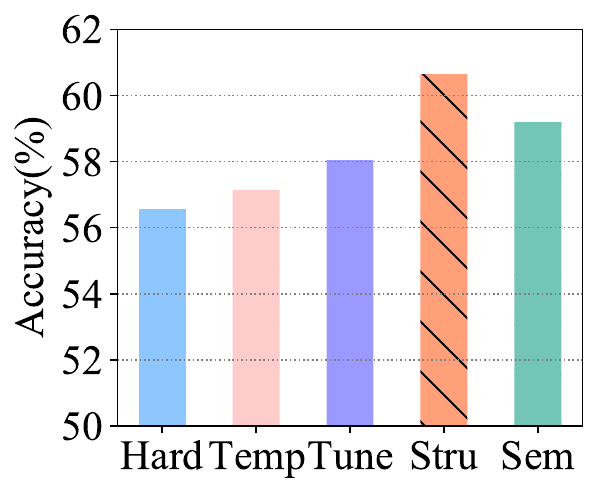}
    }
    \subfigure[Photo]{
    \includegraphics[width=0.31\linewidth]{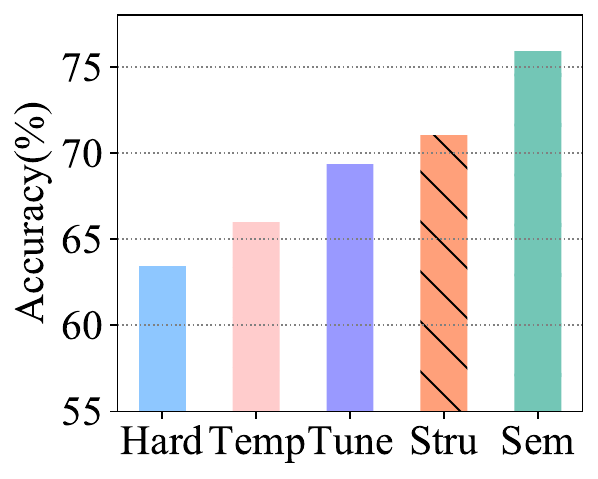}
    }
    \subfigure[Crocodile]{
    \includegraphics[width=0.31\linewidth]{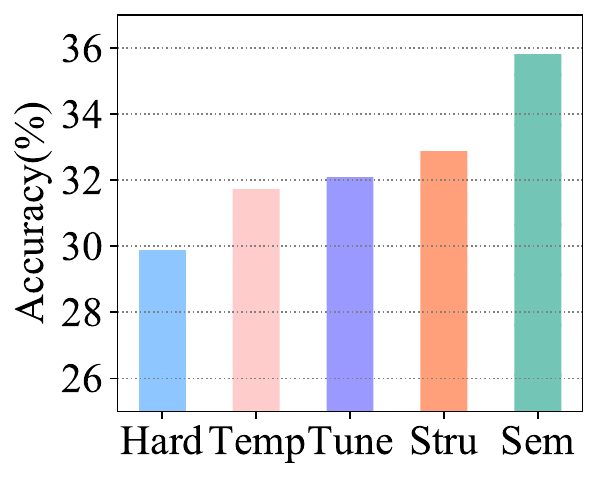}
    }
    \caption{Impact of the prompt design and different prompts in \name.}
    \label{prompts}
\end{figure}

    To analyse the effectiveness of the prompt design and different prompts in \name, we conduct the ablation study on the following variants:
    
    \noindent$\bullet$ Hard: variant without prompt and tuning.\\
    \noindent$\bullet$ Temp: variant with prompt template but without tuning.\\
    \noindent$\bullet$ Tune: variant with prompt template, tuning but without prompt injection.\\
    \noindent$\bullet$ Stru: variant with prompt template, structural prompt injection and tuning.\\
    \noindent$\bullet$ Sem: variant with prompt template, semantic prompt injection and tuning.\\
    We still follow the above setting of few-shot node classification task. In Fig.~\ref{prompts} we can see the results:\\
    (1) The degradation without prompt template mainly stems from the negative transfer problem, especially when the labeled nodes are limited.\\
    (2) The decline without tuning indicates that the projector plays a crucial role on the knowledge transfer between pre-training and downstream tasks.\\
    (3) Prompt injection improves the performance, indicating that providing extra information based on graph itself is beneficial for tasks with limited labels and the projector can effectively integrate the prompted information to enhance the performance in downstream tasks.\\ 
    (4) Structural prompt brings significant improvements on Cora while semantic prompts have better effects for Photo and Crocodile, indicating that two-hop neighbors can provide richer information on Cora, while Photo and Crocodile focus more on node attributes.
    \begin{figure}[htbp]
    \centering
    \subfigure[Cora]{
    \includegraphics[width=0.31\linewidth]{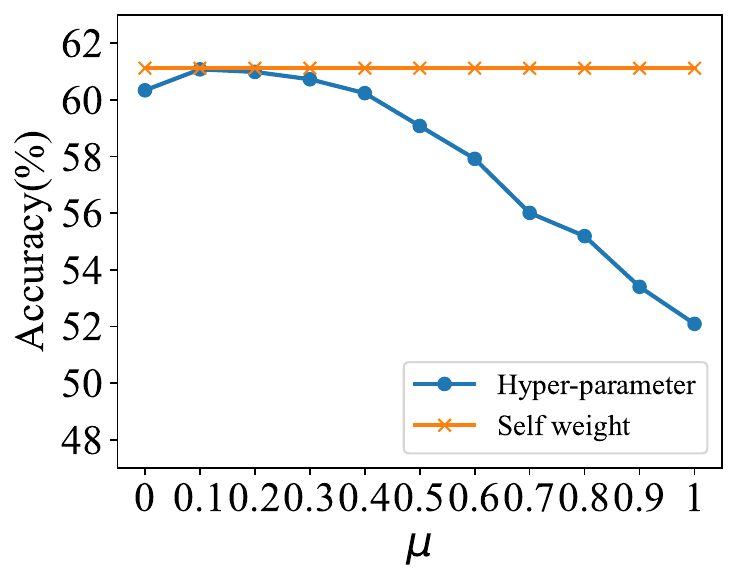}
    }
    \subfigure[Citeseer]{
    \includegraphics[width=0.31\linewidth]{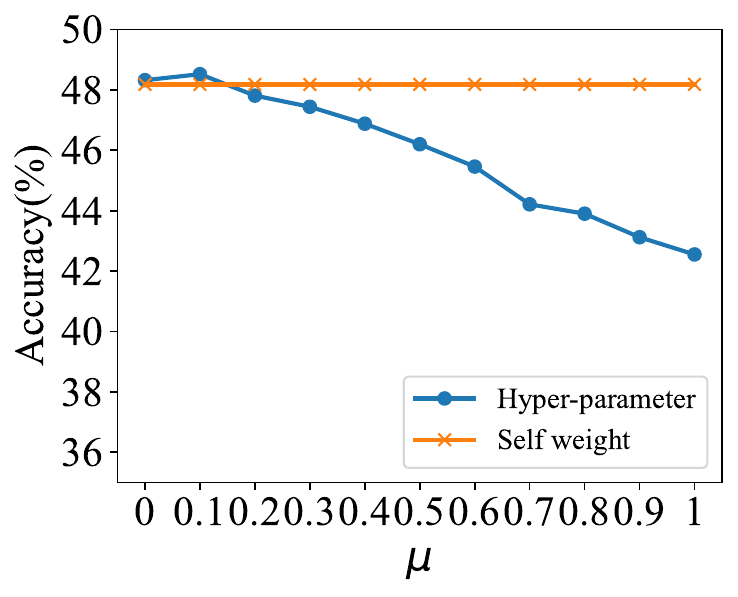}
    }
    \subfigure[Texas]{
    \includegraphics[width=0.31\linewidth]{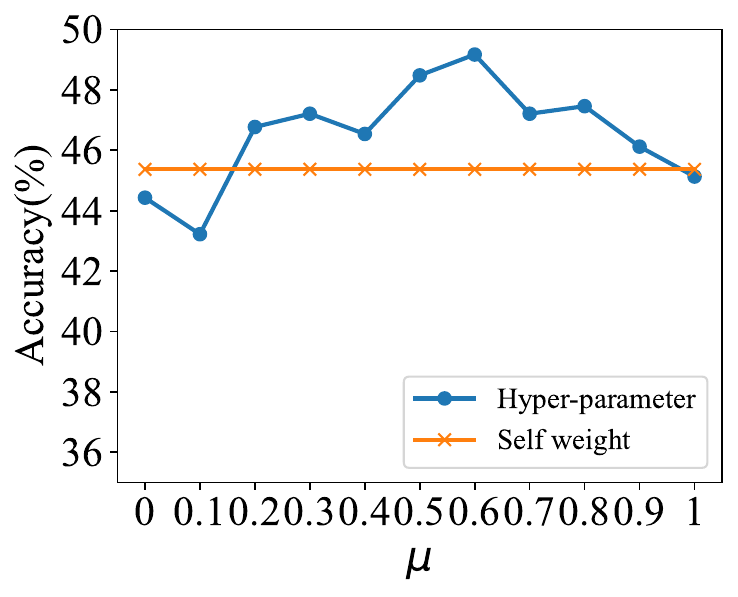}
    }
    \caption{Sensitivity analysis on the injection weight of prompts.}
    \label{parameter}
    \end{figure}
    \subsection{Parameter Analysis (RQ4)} We end this section with a sensitivity analysis on hyper-parameters in \name, i.e., the weight of prompt injection $\mu$. Meanwhile, we demonstrate the results using self weight $w$ mentioned in Section~\ref{tuning}. Specificlly, we conduct experiments on node classification by varying $\mu$ from 0 to 1 to compare with the self weight strategy. Fig.~\ref{parameter} illustrates the accuracy under different $\mu$ on Cora, Citeseer and Texas. We see that self weight strategy and smaller injection weights can assist in node classification on homophilous graph, while larger injection ratios are beneficial for heterophilous graphs where semantic information is more crucial.

\section{Conclusion}
    In this paper, we proposed \name, a framework utilizing self-prompting and tuning mechanism for graphs. The framework incorporated asymmetric graph contrastive learning to capture complex patterns for graphs with homophily and heterophily and aligned the pretext and downstream tasks into the same template. Then we reused the projector in the pre-training phase as the downstream adapter for efficient tuning and injected prompts from semantic and structure perspectives based on the information inherented in the graph itself. Finally, our experimental results demonstrated the superiority of \name.

\section{Acknowledgement}
This work is supported by National Natural Science Foundation of China No. 62202172 and Shanghai Science and Technology Committee General Program No. 22ZR1419900.

\bibliographystyle{splncs04}

\begin{thebibliography}{10}
\providecommand{\url}[1]{\texttt{#1}}
\providecommand{\urlprefix}{URL }
\providecommand{\doi}[1]{https://doi.org/#1}

\bibitem{brown2020language}
Brown, T., Mann, B., Ryder, N., Subbiah, M., Kaplan, J.D., Dhariwal, P., Neelakantan, A., Shyam, P., Sastry, G., Askell, A., et~al.: Language models are few-shot learners. Advances in neural information processing systems  \textbf{33},  1877--1901 (2020)

\bibitem{chen2022drug}
Chen, H., Lu, Y., et~al.: A drug combination prediction framework based on graph convolutional network and heterogeneous information. IEEE/ACM Transactions on Computational Biology and Bioinformatics  (2022)

\bibitem{fang2024universal}
Fang, T., Zhang, Y., Yang, Y., Wang, C., Chen, L.: Universal prompt tuning for graph neural networks. Advances in Neural Information Processing Systems  \textbf{36} (2024)

\bibitem{gong2024towards}
Gong, C., Cheng, Y., Li, X., Shan, C., Luo, S., Shi, C.: Towards learning from graphs with heterophily: Progress and future. arXiv e-prints pp. arXiv--2401 (2024)

\bibitem{guo2024architecture}
Guo, X., Wang, Y., Wei, Z., Wang, Y.: Architecture matters: Uncovering implicit mechanisms in graph contrastive learning. Advances in Neural Information Processing Systems  \textbf{36} (2024)

\bibitem{hassani2020contrastive}
Hassani, K., Khasahmadi, A.H.: Contrastive multi-view representation learning on graphs. In: International conference on machine learning. pp. 4116--4126. PMLR (2020)

\bibitem{hou2022graphmae}
Hou, Z., Liu, X., Cen, Y., Dong, Y., Yang, H., Wang, C., Tang, J.: Graphmae: Self-supervised masked graph autoencoders. In: Proceedings of the 28th ACM SIGKDD Conference on Knowledge Discovery and Data Mining. pp. 594--604 (2022)

\bibitem{hu2019strategies}
Hu, W., Liu, B., Gomes, J., Zitnik, M., Liang, P., Pande, V., Leskovec, J.: Strategies for pre-training graph neural networks. arXiv preprint arXiv:1905.12265  (2019)

\bibitem{jia2021scaling}
Jia, C., Yang, Y., Xia, Y., Chen, Y.T., Parekh, Z., Pham, H., Le, Q., Sung, Y.H., Li, Z., Duerig, T.: Scaling up visual and vision-language representation learning with noisy text supervision. In: International conference on machine learning. pp. 4904--4916. PMLR (2021)

\bibitem{jia2022visual}
Jia, M., Tang, L., Chen, B.C., Cardie, C., Belongie, S., Hariharan, B., Lim, S.N.: Visual prompt tuning. In: European Conference on Computer Vision. pp. 709--727. Springer (2022)

\bibitem{kim2023dynamic}
Kim, H., Choi, J., et~al.: Dynamic relation-attentive graph neural networks for fraud detection. arXiv preprint arXiv:2310.04171  (2023)

\bibitem{kipf2016semi}
Kipf, T.N., Welling, M.: Semi-supervised classification with graph convolutional networks. arXiv preprint arXiv:1609.02907  (2016)

\bibitem{kipf2016variational}
Kipf, T.N., Welling, M.: Variational graph auto-encoders. arXiv preprint arXiv:1611.07308  (2016)

\bibitem{li2023s}
Li, J., Wu, R., Sun, W., Chen, L., Tian, S., Zhu, L., Meng, C., Zheng, Z., Wang, W.: What's behind the mask: Understanding masked graph modeling for graph autoencoders. In: Proceedings of the 29th ACM SIGKDD Conference on Knowledge Discovery and Data Mining. pp. 1268--1279 (2023)

\bibitem{li2023seegera}
Li, X., Ye, T., Shan, C., Li, D., Gao, M.: Seegera: Self-supervised semi-implicit graph variational auto-encoders with masking. In: Proceedings of the ACM Web Conference 2023. pp. 143--153 (2023)

\bibitem{li2022finding}
Li, X., Zhu, R., Cheng, Y., Shan, C., Luo, S., Li, D., Qian, W.: Finding global homophily in graph neural networks when meeting heterophily. In: International Conference on Machine Learning. pp. 13242--13256. PMLR (2022)

\bibitem{liu2022revisiting}
Liu, N., Wang, X., Bo, D., Shi, C., Pei, J.: Revisiting graph contrastive learning from the perspective of graph spectrum. Advances in Neural Information Processing Systems  \textbf{35},  2972--2983 (2022)

\bibitem{liu2023pre}
Liu, P., Yuan, W., Fu, J., Jiang, Z., Hayashi, H., Neubig, G.: Pre-train, prompt, and predict: A systematic survey of prompting methods in natural language processing. ACM Computing Surveys  \textbf{55}(9),  1--35 (2023)

\bibitem{liu2023beyond}
Liu, Y., Zheng, Y., Zhang, D., Lee, V.C., Pan, S.: Beyond smoothing: Unsupervised graph representation learning with edge heterophily discriminating. In: Proceedings of the AAAI conference on artificial intelligence. vol.~37, pp. 4516--4524 (2023)

\bibitem{liu2023graphprompt}
Liu, Z., Yu, X., Fang, Y., Zhang, X.: Graphprompt: Unifying pre-training and downstream tasks for graph neural networks. In: Proceedings of the ACM Web Conference 2023. pp. 417--428 (2023)

\bibitem{ma2023hetgpt}
Ma, Y., Yan, N., Li, J., Mortazavi, M., Chawla, N.V.: Hetgpt: Harnessing the power of prompt tuning in pre-trained heterogeneous graph neural networks. arXiv preprint arXiv:2310.15318  (2023)

\bibitem{pei2020geom}
Pei, H., Wei, B., Chang, K.C.C., Lei, Y., Yang, B.: Geom-gcn: Geometric graph convolutional networks. arXiv preprint arXiv:2002.05287  (2020)

\bibitem{rozemberczki2021multi}
Rozemberczki, B., Allen, C., Sarkar, R.: Multi-scale attributed node embedding. Journal of Complex Networks  \textbf{9}(2),  cnab014 (2021)

\bibitem{sun2022gppt}
Sun, M., Zhou, K., He, X., Wang, Y., Wang, X.: Gppt: Graph pre-training and prompt tuning to generalize graph neural networks. In: Proceedings of the 28th ACM SIGKDD Conference on Knowledge Discovery and Data Mining. pp. 1717--1727 (2022)

\bibitem{sun2023all}
Sun, X., Cheng, H., Li, J., Liu, B., Guan, J.: All in one: Multi-task prompting for graph neural networks  (2023)

\bibitem{sun2023graph}
Sun, X., Zhang, J., Wu, X., Cheng, H., Xiong, Y., Li, J.: Graph prompt learning: A comprehensive survey and beyond. arXiv preprint arXiv:2311.16534  (2023)

\bibitem{tan2023virtual}
Tan, Z., Guo, R., Ding, K., Liu, H.: Virtual node tuning for few-shot node classification. arXiv preprint arXiv:2306.06063  (2023)

\bibitem{velivckovic2017graph}
Veli{\v{c}}kovi{\'c}, P., Cucurull, G., Casanova, A., Romero, A., Lio, P., Bengio, Y.: Graph attention networks. arXiv preprint arXiv:1710.10903  (2017)

\bibitem{velivckovic2018deep}
Veli{\v{c}}kovi{\'c}, P., Fedus, W., Hamilton, W.L., Li{\`o}, P., Bengio, Y., Hjelm, R.D.: Deep graph infomax. arXiv preprint arXiv:1809.10341  (2018)

\bibitem{wang2021afec}
Wang, L., Zhang, M., Jia, Z., Li, Q., Bao, C., Ma, K., Zhu, J., Zhong, Y.: Afec: Active forgetting of negative transfer in continual learning. Advances in Neural Information Processing Systems  \textbf{34},  22379--22391 (2021)

\bibitem{wei2021finetuned}
Wei, J., Bosma, M., Zhao, V.Y., Guu, K., Yu, A.W., Lester, B., Du, N., Dai, A.M., Le, Q.V.: Finetuned language models are zero-shot learners. arXiv preprint arXiv:2109.01652  (2021)

\bibitem{wu2023heterophily}
Wu, Q., Yang, Y., et~al.: Heterophily-aware social bot detection with supervised contrastive learning. arXiv preprint arXiv:2306.07478  (2023)

\bibitem{xia2022survey}
Xia, J., Zhu, Y., Du, Y., Li, S.Z.: A survey of pretraining on graphs: Taxonomy, methods, and applications. arXiv preprint arXiv:2202.07893  (2022)

\bibitem{xiao2024simple}
Xiao, T., Zhu, H., Chen, Z., Wang, S.: Simple and asymmetric graph contrastive learning without augmentations. Advances in Neural Information Processing Systems  \textbf{36} (2024)

\bibitem{yuan2023muse}
Yuan, M., Chen, M., Li, X.: Muse: Multi-view contrastive learning for heterophilic graphs. arXiv preprint arXiv:2307.16026  (2023)

\bibitem{zhang2022localized}
Zhang, H., Wu, Q., Wang, Y., Zhang, S., Yan, J., Yu, P.S.: Localized contrastive learning on graphs. arXiv preprint arXiv:2212.04604  (2022)

\bibitem{zhao2023graphglow}
Zhao, W., Wu, Q., Yang, C., Yan, J.: Graphglow: Universal and generalizable structure learning for graph neural networks. In: Proceedings of the 29th ACM SIGKDD Conference on Knowledge Discovery and Data Mining. pp. 3525--3536 (2023)

\bibitem{zhu2020beyond}
Zhu, J., Yan, Y., Zhao, L., Heimann, M., Akoglu, L., Koutra, D.: Beyond homophily in graph neural networks: Current limitations and effective designs. Advances in neural information processing systems  \textbf{33},  7793--7804 (2020)

\bibitem{zhu2020deep}
Zhu, Y., Xu, Y., Yu, F., Liu, Q., Wu, S., Wang, L.: Deep graph contrastive representation learning. arXiv preprint arXiv:2006.04131  (2020)

\end{thebibliography}

\end{document}